\documentclass{article}
\usepackage{spconf, amsmath, amsfonts, amssymb,graphicx}
\usepackage{enumitem}
\usepackage{multirow}
\usepackage{booktabs}
\usepackage{ifluatex}
\usepackage{svg}
\usepackage[redeflists]{IEEEtrantools}
\usepackage{hyperref}
\hypersetup{hidelinks}
\usepackage{xurl}
\setlist{nosep, leftmargin=14pt}

\usepackage{mwe} % to get dummy images

\title{Learning to Estimate Critical Gait Parameters from Single-View RGB Videos with Transformer-Based Attention Network}
\name{Hung Le, Hieu Pham$^{\star}$ \thanks{$^{\star}$Corresponding author: \textcolor{blue}{hieu.ph@vinuni.edu.vn}}}
\address{College of Engineering \& Computer Science, VinUni-Illinois Smart Health Center, VinUniversity}

\begin{document}

%\ninept
%
\maketitle
\begin{abstract}
Musculoskeletal diseases and cognitive impairments in patients lead to difficulties in movement as well as negative effects on their psychological health. Clinical gait analysis, a vital tool for early diagnosis and treatment, traditionally relies on expensive optical motion capture systems. Recent advances in computer vision and deep learning have opened the door to more accessible and cost-effective alternatives. This paper introduces a novel spatio-temporal Transformer network to estimate critical gait parameters from RGB videos captured by a single-view camera. Empirical evaluations on a public dataset of cerebral palsy patients indicate that the proposed framework surpasses current state-of-the-art approaches and show significant improvements in predicting general gait parameters (including Walking Speed, Gait Deviation Index - GDI, and Knee Flexion Angle at Maximum Extension), while utilizing fewer parameters and alleviating the need for manual feature extraction.
\end{abstract}
\section{Introduction}
\label{sec:intro}

Approximately 1.7 billion individuals worldwide are currently affected by movement-related diseases \cite{cieza2020global}. These conditions encompass a wide range of diseases that negatively impact the muscles, bones, and joints. The cognitive function of a person is also shown to be associated with their manner of movement \cite{montero2012gait}. As a result, human gait analysis has become a crucial tool for early diagnosis of musculoskeletal conditions. In clinical gait analysis, quantitatively measuring general gait parameters, such as walking speed, cadence, or gait deviation index (GDI) is essential in determining the overall health and function of a patient’s gait \cite{schwartz2008gait, brandstater1983hemiplegic,guffey2016gait}.

The current workflow of human motion analysis typically relies on optical motion capture systems (Mocap) \cite{pfister2014comparative}. These systems require the use of multiple specialized cameras and sensors to track the patient's movements. The two-dimensional (2D) videos from these cameras are then constructed into three-dimensional (3D) motion data \cite{cheung2000real}, which then can be used to compute various gait parameters. Unfortunately, Mocap systems require expensive equipment and well-trained personnel while also demanding careful calibration and positioning of the system for optimal performance \cite{merriaux2017study}. Alternatives include wearable inertial sensors or depth cameras \cite{picerno201725, xu2015accuracy} but their accessibility is limited due to their cost and the specialized skills required for operation.

Recently, 2D markerless video sequences captured from single-camera videos have emerged as a convenient and economical option \cite{bauckhage2009automatic} by minimizing the needs for expensive equipment. In particular, pose estimation models become a powerful tool to automatically learn and extract 2D skeletal data \cite{cao2017realtime} for human gait assessment \cite{stenum2021two}. However, estimating gait parameters directly from these skeletal data is prone to errors due to the lack of robustness and precision from pose tracking algorithms \cite{seethapathi2019movement}. As a result, the expressiveness and resilience to noise of deep neural networks enable them to be a more efficient and precise method for estimating general gait parameters \cite{kidzinski2020deep}. In this study, we propose a novel spatio-temporal Transformer architecture for gait analysis using single-camera RGB videos, which could effectively learn intricate spatio-temporal dynamics in human gait, a challenge for most traditional methods. The first stage of this architecture aims to extract the keypoints of 2D human pose captured from a video using OpenPose \cite{cao2017realtime}. The second stage uses a spatio-temporal Transformer network to predict gait parameters, including Gait Deviation Index, knee flexion angle at maximum extension, speed, and cadence. The main contributions of this work are summarized as follows: 

\begin{itemize}
    \item We are the first to introduce a spatio-temporal Transformer network for quantitatively estimating gait parameters for gait analysis from single-view RGB videos. Our network design allows us to extract both spatial and temporal features effectively from the anatomical keypoints taken from single-camera RGB videos.
    \item The flexibility inherent in the proposed network helps to achieve superior performance in predicting gait parameters compared to current state-of-the-art methods, meanwhile reducing number of parameters and obviating the need for manually crafted time series.
    \item Our codes, and trained deep learning models are made publicly available: \color{blue}{\url{https://github.com/vinuni-vishc/transformer-gait-analysis}}. 

\end{itemize}
\section{Related Works}

Estimating gait parameters directly from single-camera RGB videos has been proposed in previous studies \cite{kidzinski2020deep,azhand2021algorithm,lonini2022video,cotton2022transforming}. For instance, Azhand \textit{et al.} \cite{azhand2021algorithm} extracted both 2D and 3D skeleton from a 2D input video and optimized the 3D pose based on patient’s heights to quantify gait parameters. Following the success of CNN in vision tasks, Kidzinski \textit{et al.} \cite{kidzinski2020deep} proposed to use OpenPose \cite{cao2017realtime} for keypoint extraction and a one-dimensional convolutional neural network (1D-CNN) model to predict clinically relevant motion parameters from an ordinary video of a patient. Similarly, Lonini \textit{et al.} \cite{lonini2022video} explored 1D-CNN to estimate gait parameters while utilizing DeepLabCut \cite{mathis2018deeplabcut} for tracking anatomical keypoints of stroke patients. Researchers also employ Spatial-Temporal Graph Convolutional Network (STGCN) for gait parameter prediction \cite{jalata2021movement}. For the graph of skeletal sequences, the connections of the joints are based on anatomical structure while every joint is connected to itself in the previous and the next frame. Transformer architecture is also proposed to compute interpretable kinematic parameters from estimated 3D pose, which are then used to estimate patients' gait cycle and produce final gait parameters \cite{cotton2022transforming}.

\section{Methodology}
\subsection{Problem Formulation}
Given an input sequence of motion $\mathbf{X} \in \mathbb{R}^{T \times N \times 2}$, it is defined as $ \mathbf{X} = [\mathbf{x}_1, \mathbf{x}_2, ..., \mathbf{x}_T]$ where $T$ is the number of frames in a video segment. Each frame $\mathbf{x}_i = [\mathbf{j}_i^{(1)}, \mathbf{j}_i^{(2)},..., \mathbf{j}_i^{(N)}]$ denotes a human pose at time step $i$ with $N$ joints where joint $\mathbf{j}_i^{(n)}$ is a 2D vector, representing the $(x,y)$ coordinates of the joint in the Cartesian plane. 

The output $\mathbf{y} \in \mathbb{R}^{1}$ is the gait parameter for a patient. We formulate this problem as a fully supervised learning task where we aim to learn a non-linear mapping $f_{\theta}: \mathbb{R}^{T \times N \times 2} \rightarrow \mathbb{R}^1$, here $f_{\theta}$ parameterized by $\theta$. Given a dataset  $\mathcal{D} = \{(\mathbf{X_k}, \mathbf{y_k}): \mathbf{X}_k \in \mathbb{R}^{T \times N \times 2}, \mathbf{y}_k \in  \mathbb{R}^{1}, k = 1, 2, ..., S \}$ with $S$ samples, we learn the optimal parameters $\theta^*$ by minimizing the loss function $\mathcal{L}$ over the dataset $\mathcal{D}$ as
\begin{equation}
\theta^* = \underset{\theta}{\arg\min}\sum_{k=1}^S\mathcal{L}(f_{\theta}(\mathbf{X}_k), \mathbf{y}_k). \nonumber
\end{equation}

\begin{figure*}[t]
\fontsize{6}{7}\selectfont
\centering
{\includegraphics[width = 0.947\linewidth, keepaspectratio]{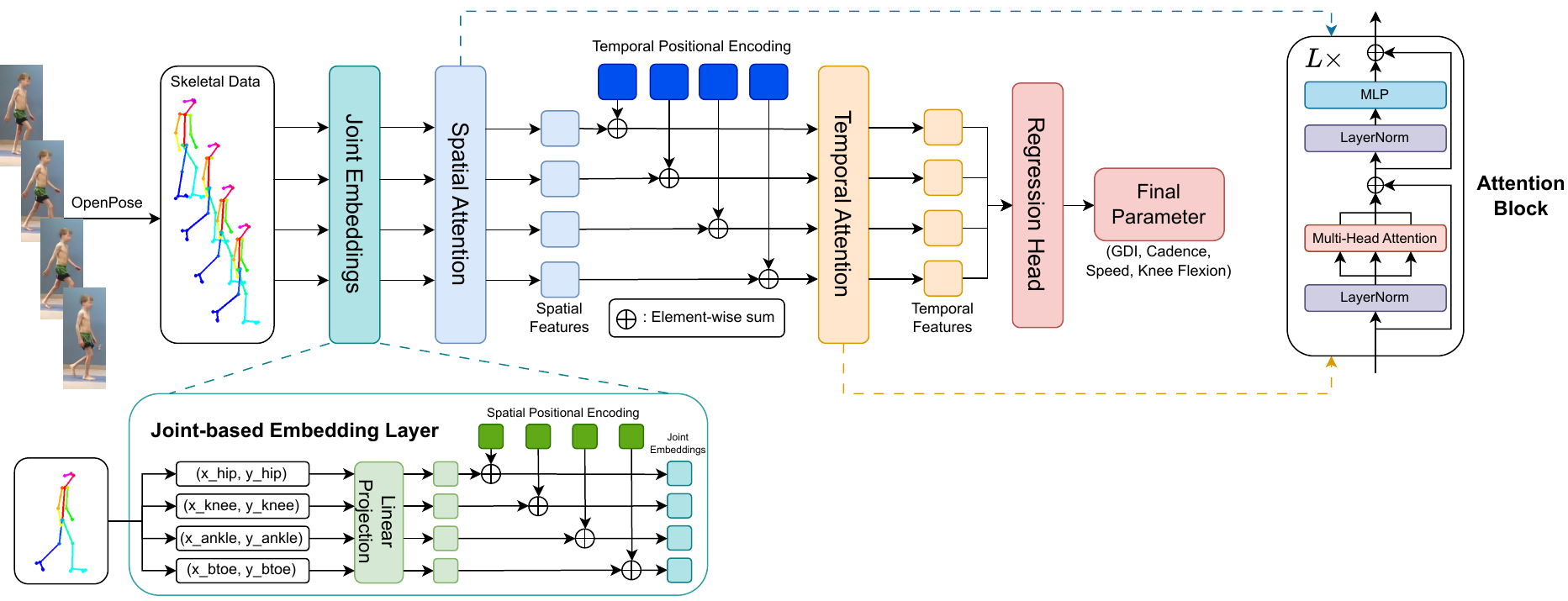}}
\caption{Overview of our approach for $T = 4$. We first project the 2D coordinates of each joint to a $D$-dimensional space. Our architecture has two attention blocks: spatial and temporal attention block, adopting multi-head attention from \cite{vaswani2017attention} \cite{dosovitskiy2020image}. The spatial attention block extracts spatial information by attending to every other joint in the same frame. The temporal attention block captures temporal dependencies among the frames given a motion sequence. Lastly, we use a Fully Connected Neural Network to output the final parameters.}
\label{fig:1}
\end{figure*}

\subsection{Network Architecture}
We designed a Transformer-based network to estimate critical gait parameters directly from single-camera RGB videos. The network comprises of two main blocks: spatial attention and temporal attention blocks. This architecture is motivated by the success of the architecture proposed in \cite{zheng20213d} for three-dimensional pose estimation. The key idea behind self-attention is it enables the network to weigh the importance of every other element in a sequence when encoding each element, taking into account the relationships between them. In the spatial attention block, it determines the amount of focus of one particular joint has on every other joint in a single frame. This allows the model to learn the necessary spatial information between the joints.

For every frame, we embed each joint into high dimensional feature by projecting the joint's coordinates in two-dimensional space to $D$-dimensional vector via a linear transformer layer. It is formulated as
\begin{equation}
\mathbf{z}_s = [\mathbf{x}_1W_{s} + E_s, \mathbf{x}_2W_{s} + E_s, ..., \mathbf{x}_TW_{s} + E_s].\label{eq1}
\end{equation}

At time step $i$, the $D$-dimensional embedding $\mathbf{z}_{s,i}$ of an input frame $\mathbf{x}_i \in \mathbb{R}^{1 \times N \times 2}$ is computed with a projection matrix $W_{s} \in \mathbb{R}^{2 \times D}$. To retain the spatial relations between the individual joint, we add a learnable spatial positional encoding term $E_{s} \in \mathbb{R}^{1 \times N \times D}$ to the embedding. The joint embeddings $\mathbf{z}_s \in \mathbb{R}^{T \times N \times D}$ are then passed to the spatial attention block.

Following the scaled dot-product attention \cite{vaswani2017attention}, we project the joint embeddings to query $\mathbf{Q}$, key $\mathbf{K}$ and value $\mathbf{V}$ matrices with learnable weight matrices $W^{\mathbf{Q}}$, $W^{\mathbf{K}}$, $W^{\mathbf{V}}$. For notation clarity, $\mathbf{Q}_{s}, \mathbf{K}_{s}, \mathbf{V}_{s}$ denote the query, key and value matrices for the spatial attention block, which include
\begin{equation}\label{eq2}
\mathbf{Q}_{s} = \mathbf{z}_{s}W^{\mathbf{Q}}_s, \quad \mathbf{K}_{s} = \mathbf{z}_{s}W^{\mathbf{K}}_s, \quad \mathbf{V}_{s} = \mathbf{z}_{s}W^{\mathbf{V}}_s
\end{equation}

For each attention block, we use multi-head attention (MHA) mechanism. In a multi-head attention block, multiple attention heads are calculated in parallel and independently. The output of this layer will be the concatenation of the outputs of the $H$ heads
\begin{IEEEeqnarray}{rCl}
\text{Attention}(\mathbf{Q}, \mathbf{K}, \mathbf{V}) &=& \text{Softmax}(\mathbf{Q}\mathbf{K}^T/\sqrt{D})\mathbf{V}\label{eq3}\\
\text{MHA}(\mathbf{Q}, \mathbf{K}, \mathbf{V}) &=& \text{Concat}(\text{head}_1, ..., \text{head}_H)W^{out}\label{eq4}\\
\text{head}_h &=& \text{Attention}(\mathbf{Q}_h, \mathbf{K}_h, \mathbf{V}_h),\label{eq5}
\end{IEEEeqnarray}
for $h \in [1, 2, ..., H]$. Given our initial embedding $\mathbf{z}_{s}$ or $\mathbf{z}_s^{(0)}\in \mathbb{R}^{T \times N \times D}$, we feed it into the spatial attention block (as illustrated in Fig. \ref{fig:1}) with $L$ layers
\begin{IEEEeqnarray}{rCl}
\mathbf{z}_s^{(\ell)'} &=& \text{MHA}(\text{LayerNorm}(\mathbf{z}_{s}^{(\ell - 1)})) + \mathbf{z}_{s}^{(\ell - 1)}\label{eq6}\\
\mathbf{z}_s^{(\ell)} &=& \text{MLP}(\text{LayerNorm}(\mathbf{z}_{s}^{(\ell - 1)})) + {z}_s^{(\ell)'}\label{eq7}\\
\mathbf{z}_s^{out} &=& \text{LayerNorm}(\mathbf{z}_{s}^{(L)})\label{eq8},
\end{IEEEeqnarray}
where $\ell \in [1, 2, ..., L]$, LayerNorm denotes layer normalization and MLP denotes Multi-Layer Perceptron. The output of the spatial attention block $\mathbf{z}_s^{out} \in \mathbb{R}^{T \times N \times D}$ is then flattened to $\mathbf{z}_t^{(0)} \in \mathbb{R}^{T \times (N \cdot D)}$, being the input to the temporal attention block. 

Similar to the spatial attention block, we add a learnable temporal positional embedding term $E_t \in \mathbb{R}^{T \times (N \cdot D)}$ to encode the position of each frame and compute $\mathbf{Q}_t, \mathbf{K}_t, \mathbf{V}_t$ for each frame. Following the same architecture of the spatial attention block in \eqref{eq6}, \eqref{eq7}, \eqref{eq8}, the output of the temporal attention block is $\mathbf{z}_t^{out} \in \mathbb{R}^{T \times (N \cdot D)}$. Since flattening $\mathbf{z}_t^{out}$ and feeding it to a Fully Connected Neural Network (FCNN) in a similar way to most modern CNN architectures could vastly increase the total number of parameters in the model. Therefore, we take a weighted sum of $\mathbf{z}_t^{out}$ by the frame dimension to produce $\hat{\mathbf{y}}' \in \mathbb{R}^{1 \times (N \cdot D)}$. Finally, pass $\hat{\mathbf{y}}'$ to a FCNN to produce the final prediction. Mean Squared Error (MSE) is used to minimize the error between the ground truth parameter and the predicted parameter.
\section{Experiments}
\subsection{Datasets and Experimental Settings}
We validated our approach on a publicly available dataset collected at Gillette Children’s Specialty Healthcare between 1994 and 2015. Following the preprocessing steps described in \cite{kidzinski2020deep}, the dataset consists of 2,212 videos of 1,138 patients, who had been diagnosed with cerebral palsy. The average age of the patients is 13 years old with standard deviation of 6.9 years old. The average height and weight of the patients are 141cm and 41kg with standard deviation of 19cm and 17kg, respectively. The raw videos were initially captured at 1280$\times$960 resolution and then downsampled to 640$\times$480 in order to better accommodate the capability of most low-end cameras. Each video is then divided into multiple 124-frame segments by window slicing with an offset of 31 frames. For experiments, the dataset is split into training, validation, and test set with the ratio of 8:1:1. It is divided by patients' ID to ensure no patient is included in different set. The training set consists of 1,768 videos of 920 patients, meanwhile the validation set consists of 212 videos of 106 patients, and the test set consists of 232 videos of 112 patients.

\subsection{Implementation Details} In our experiments, the length of one video segment (in frames) is $T = 124$, and the spatial embedding dimension $D = 12$. We only use four joints directly related to kinematic characteristics of the gait, which are hip, knee, ankle and big toe. For GDI and knee flexion angle at maximum extension, we predict the parameter for each side of the gait while speed and cadence are measured using both sides of the gait.

For both spatial and temporal attention blocks, the number of layers is $L = 1$ and the number of head is $H = 2$. The model is optimized with Adam for 200 epochs with batch size of 128. A cosine annealing scheduler with warm restarts was used as proposed in \cite{loshchilov2016sgdr} with $T_0 = 40, T_{mult} = 1$. For speed and cadence, the initial learning rate is $6e-4$ and the minimum learning rate is $1e-4$. For GDI and knee flexion angle, the initial learning rate is $3e-4$ and the minimum learning rate is $8e-5$. The best weights are selected based on the validation error. We implemented the proposed model in PyTorch and trained it on only one NVIDIA GPU 3080. 
\def\arraystretch{1.8}% 
\begin{table}[t]
  \centering
  \caption{\label{tab1}Quantitative Assessment of the proposed model in Estimating Gait Parameters. Best results are in bold.}
  \resizebox{\columnwidth}{!}{%
  \begin{tabular}[c]{l|c|c|c|c|c|c}
    \toprule
    \multirow{2}{*}{\textbf{Gait Parameters}} & \multicolumn{3}{c|}{\textbf{Correlation}}                                                  & \multicolumn{3}{c}{\textbf{Mean Absolute Error}} \\ 
    \cline{2-7} & CNN & STT (ours) & $\Delta \uparrow$ & CNN & STT (ours) & $\Delta \downarrow$\\
    \hline\hline
    GDI  &  0.7379  &   \textbf{0.7466}   & 1.18\% &  6.5469 & \textbf{6.3137}  & 3.56\% \\
    Knee Flexion (degrees)  &  0.8298  &  \textbf{0.8384} & 1.04\% & 5.9129 &  \textbf{5.8220} & 1.54\%\\
    Speed (m/s)  &  0.7519  &   \textbf{0.7565} &  0.62\%  & 0.1482 & \textbf{0.1481} & 0.13\%\\
    Cadence (strides/s)  &  \textbf{0.7601}  &  0.7421 & -2.37\% & \textbf{0.1035} &  0.1078  & -4.12\%\\
    \bottomrule
    \end{tabular}%
    }
\end{table}

\subsection{Experimental Results}

To quantify the effectiveness of our model, we compared our network with current state-of-the-art model (1D-CNN) \cite{kidzinski2020deep} on the test set. The performance of the proposed method is measured using the degree of correlation and mean absolute error (MAE). The predicted values of our Spatio-Temporal Transformer for GDI, knee flexion angle at maximum extension, walking speed and cadence achieved $0.75, 0.84, 0.76, 0.74$ correlation, respectively with ground-truth parameters. Table \ref{tab1} shows that the proposed approach achieves better performance, approximately $1\%$ for correlation for three (GDI, Knee Flexion, Speed) out of four parameters while having $36.5\%$ fewer parameters than 1D-CNN. Our method also yields a $3.5\%$ and $1.5\%$ decrease in MAE for GDI and knee flexion angle. In addition, when 1D-CNN utilized both skeletal data and hand-engineered time series to predict outcomes, we leveraged only skeletal motion sequence to estimate the final parameters.

\begin{figure}
    \centering
    \includegraphics[width = 0.95\columnwidth]{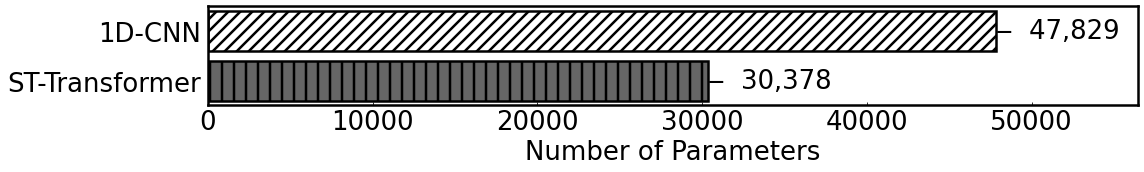}
    \caption{Comparison of the number of parameters of 1D-CNN and the proposed Spatio-Temporal Transformer network.}
    \label{fig:2}
\end{figure}

\begin{figure}[t]
    \centering
    \includegraphics[width = 0.9\columnwidth]{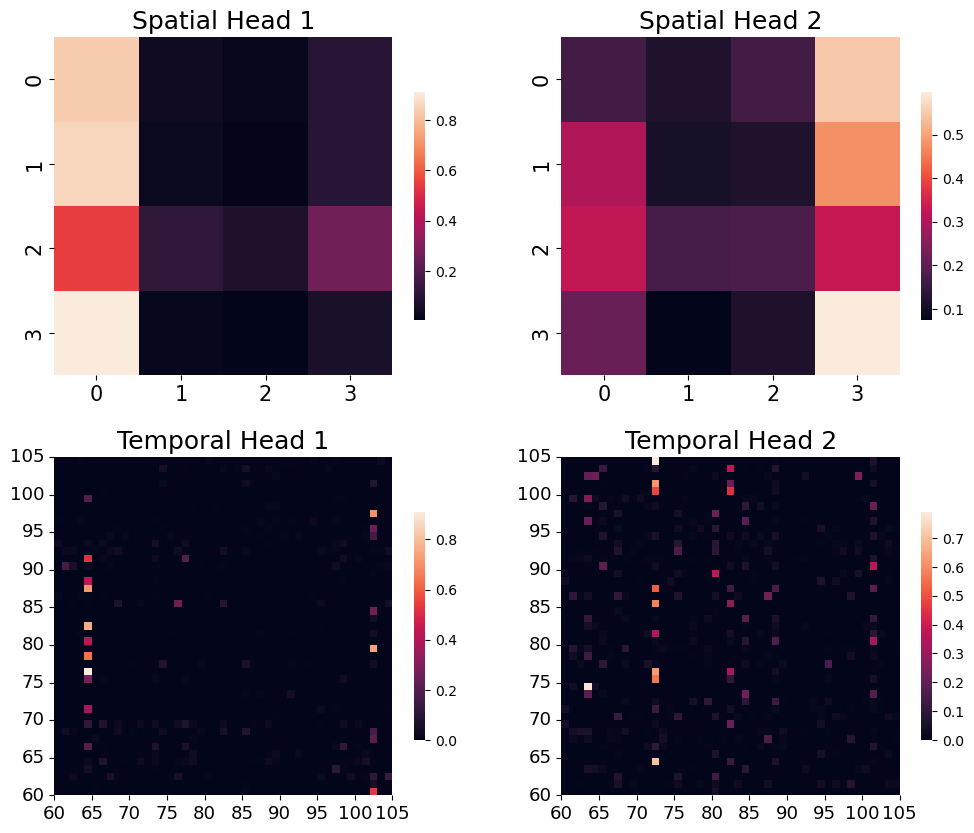}
    \caption{Visualizing the attention matrix $\mathbf{A}$ of different heads in the spatial attention block at timestep $t=46$ (top) and temporal attention block from timestep $t=60$ to $t=105$ (bottom) for GDI measurement. Entry $\mathbf{A}_{i,j}$ denotes the attention weight between joint $i$ and joint $j$ or frame $i$ and frame $j$}
    \label{fig:3}
\end{figure}

\section{Discussion and Conclusion}
Intuitively, the 1D-CNN architecture only captures the temporal dependency of each joint or each time series independently. As a result, it fails to gather the underlying spatial information of the joints and needs to rely on hand-crafted features. On other the hand, the proposed spatio-temporal Transformer can be conceptualized as a graph neural network, characterized by a fully connected graph structure. Spatial and temporal connections within this architecture are established through attention scores derived from the multi-head attention block. This design facilitates enhanced flexibility for extracting motion features from single-camera videos. The inherent capacity of our proposed architecture facilitates effective generalization of gait parameters from anatomical keypoints, eliminating the necessity for manual curation of gait features.

Analyzing self-attention mechanism, Fig. \ref{fig:3} visualizes the attention score of the spatial and temporal attention block for GDI prediction. For the spatial attention block, there is a strong focus on joint 0 (hip) for Head 1 while Head 2 concentrates on joint 3 (big toe). For temporal attention, Head 1 learns the dependencies among frame 65 with frames 82 and 87 while the connections of frame 72 with frames 64, 77, 86, and 101 are captured by Head 2. This demonstrates our approach could exploit long-term dependencies within the spatial-temporal data that most deep learning models do not have the capacity of. 

However, it is evident that our model struggles with cadence predictions. This could be due to the fact that the temporal dependencies of our model rely on attention of the frame level. As a result, it is lacking in the capacity to recognize gait cycles and strides, which are the key determinants of cadence.

To conclude, we proposed a novel spatio-temporal Transformer architecture to extract spatial and temporal features from a single-view 2D motion sequence. Experiments show that our approach achieved superior performance with a lower level of complexity in comparison to prior art. This study suggests that deep learning-based methods provide reliable and accurate evaluation for gait analysis. Our future directions are to incorporate gait events and attention-based networks to further lower prediction errors.

\section{Acknowledgments}
No funding was received for conducting this study. The authors have no relevant financial or non-financial interests to disclose.

\section{Compliance with Ethical Standards}
This research study was conducted retrospectively using human subject data made available in open access by \cite{kidzinski2020deep}. Ethical approval was not required as confirmed by the license attached with the open access data.


\small{
\bibliographystyle{IEEEbib}
\bibliography{bibliography}}

\end{document}